\documentclass[letterpaper]{article} 
\usepackage[preprint]{aaai2027}  
\usepackage[hyphens]{url}  
\usepackage{graphicx} 
\usepackage{natbib}  
\usepackage{caption} 
\usepackage{algorithm}
\usepackage{algorithmic}
\usepackage{amsmath}
\usepackage{amssymb}
\usepackage{xcolor}
\usepackage{booktabs}
\usepackage{multirow}
\usepackage{array}
\usepackage{tabularx}
\usepackage{colortbl}

\definecolor{secondblue}{HTML}{10BDF4}
\newcolumntype{Y}{>{\centering\arraybackslash}X}

\usepackage{newfloat}
\usepackage{listings}
\DeclareCaptionStyle{ruled}{labelfont=normalfont,labelsep=colon,strut=off} 
\floatstyle{ruled}
\newfloat{listing}{tb}{lst}{}
\floatname{listing}{Listing}

\usepackage{booktabs}

\title{Newton Deep Unfolding for Compressed Sensing}
\author {
   Changhua He,
    Xianchao Xiu\corresponding
}
\affiliations {
    School of Mechatronic Engineering and Automation, Shanghai University\\
    \texttt{hechanghua@shu.edu.cn}, \texttt{xcxiu@shu.edu.cn}
}

\begin{document}

\maketitle

\begin{abstract}
Compressed sensing (CS) reconstructs images from highly limited measurements, but existing deep unfolding methods are typically driven by first-order optimization and weakly exploit the optimization states generated during reconstruction. To address these limitations, we propose a Newton deep unfolding network (NDU-Net), which, to the best of our knowledge, is the first deep unfolding framework that leverages second-order optimization for CS reconstruction. Specifically, NDU-Net introduces a Newton update (NU) module to estimate Newton-type update directions and generate optimization states that characterize the current reconstruction process. Furthermore, a Newton-guided multi-scale prior (MP) module is designed to incorporate these optimization states into multi-scale feature restoration, thereby enabling the learned prior to adapt to the current reconstruction stage. Experimental results under different CS ratios confirm that our proposed NDU-Net achieves promising reconstruction performance and exhibits enhanced robustness. Our code is available at \url{https://github.com/xianchaoxiu/DNU-Net}.
\end{abstract}


\section{Introduction}

As a general signal acquisition paradigm, compressed sensing (CS) reduces the number of required measurements by exploiting the sparsity or compressibility of natural signals~\cite{donoho2006compressed,candes2006robust}. This measurement-efficient property can lower acquisition, storage, and transmission costs, making CS useful for a wide range of applications, including single-pixel imaging, magnetic resonance imaging, computed tomography, and snapshot compressive imaging~\cite{duarte2008single,lustig2007sparse,sidky2008image,yuan2021snapshot,liu2024compressive,zhao2024untrained}. With the development of deep learning, CS reconstruction has evolved from hand-crafted priors and iterative solvers toward data-driven and unfolding-based reconstruction networks~\cite{zhao2026survey}.

In this paper, we focus on CS reconstruction for images, where the goal is to recover a high-quality image from the undersampled measurement.
Mathematically, the observation model can be formulated as
\begin{equation}\label{ober}
    \mathbf{y} = \mathbf{A}\mathbf{x} + \mathbf{e},
\end{equation}
where $\mathbf{x} \in \mathbb{R}^{N}$ is the unknown image, $\mathbf{y} \in \mathbb{R}^{M}$ denotes the compressed measurement, $\mathbf{A} \in \mathbb{R}^{M \times N}$ denotes the sensing matrix, and $\mathbf{e}$ represents the measurement noise. Denote 
\begin{equation}
	f(\mathbf{x})
	=
	\frac{1}{2}
	\|
	\mathbf{A}\mathbf{x}
	-
	\mathbf{y}
	\|_{2}^{2},
	\label{eq:data_term}
\end{equation}
which measures the mismatch between the reconstructed image and the compressed measurements under the sensing matrix $\mathbf{A}$ and is often called the data-consistency term. Since $M \ll N$ in typical CS settings, recovering $\mathbf{x}$ from the observation model in Eq.~\eqref{ober} is highly ill-posed and may admit multiple plausible solutions. Therefore, accurate reconstruction requires both data consistency and effective image priors, especially at low sampling ratios, where fine textures and edges are difficult to recover.

\begin{figure}[t]
    \centering
    \includegraphics[width=\columnwidth]{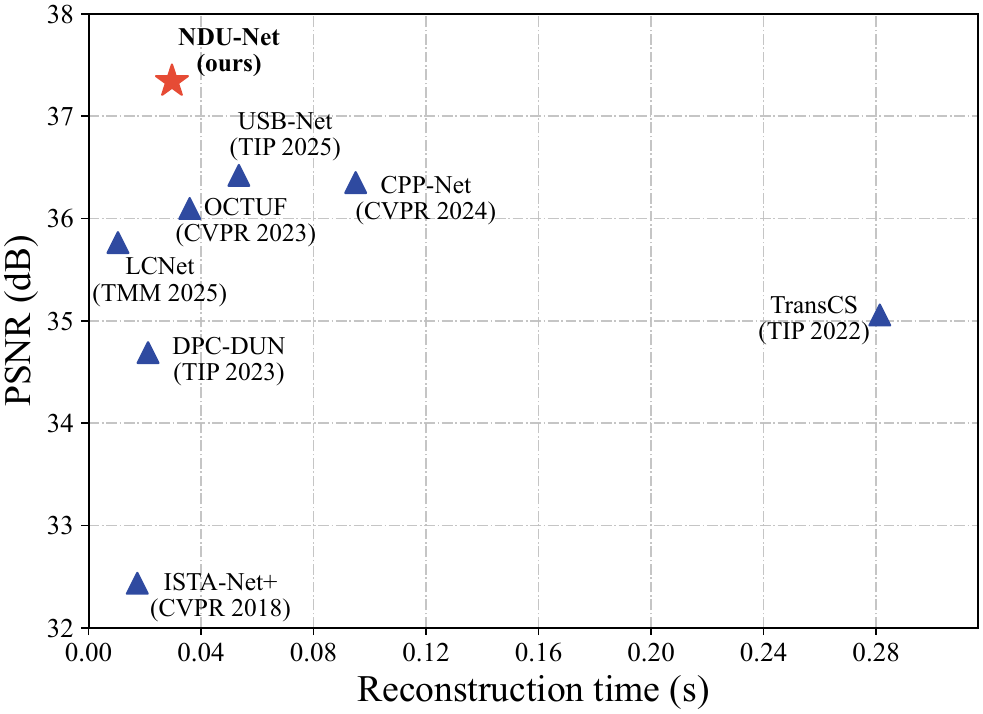}
    \caption{PSNR versus reconstruction time on Set11 at CS ratio $=0.25$.}
    \label{fig:set11_time_scatter}
\end{figure}

Generally speaking, CS reconstruction can be characterized as the following regularized problem
\begin{equation}
    \min_{\mathbf{x}}~ f(\mathbf{x}) + \lambda \mathcal{R}(\mathbf{x}),
    \label{eq:cs_regularized_model}
\end{equation}
where  $\mathcal{R}(\mathbf{x})$ describes the prior knowledge with parameter $\lambda$.
Traditional optimization-based methods solve problem~\eqref{eq:cs_regularized_model} using sparsity-promoting penalties, such as $\ell_0$, $\ell_1$, and $\ell_{1/2}$ penalties, together with iterative solvers such as ISTA, AMP, and ADMM~\cite{donoho2006compressed,candes2006robust,lustig2007sparse,sidky2008image}. These methods are usually interpretable due to the fact that each iteration is explicitly derived from an optimization objective and the roles of the data-consistency and prior terms are clearly defined. However, their reconstruction performance heavily depends on the choice of $\mathcal{R}(\mathbf{x})$, $\lambda$, and solvers. Besides, iterative optimization often leads to high computational cost. In contrast, deep neural network-based methods improve CS reconstruction by learning image priors directly from training data~\cite{kulkarni2016reconnet,shi2017deepnetworks,shi2019scsnet}. Compared with hand-crafted priors, learned priors can better capture complex image structures and usually achieve higher reconstruction quality with fast feed-forward inference. Nevertheless, these methods often lack an explicit connection to the physical measurement model, which may weaken data consistency and reduce interpretability when facing low sampling ratios or noisy measurements.

Deep unfolding networks bridge iterative optimization and neural network design by unfolding optimization algorithms into a sequence of trainable stages. For CS reconstruction, representative methods include ISTA-Net, ISTA-Net++, AMP-Net, LDAMP, and ADMM-Net~\cite{zhang2018istanet,you2021istanetpp,zhang2021ampnet,metzler2017ldamp,yang2016deepadmm}. Despite the progress of existing CS unfolding methods~\cite{shen2025hunet,cui2026mhcdun}, all of them still perform the data-consistency update through first-order correction or simple learned variants. In addition, the optimization states produced during this update, such as gradients, residuals, and update directions, are not sufficiently considered by the learned prior module. \textit{Thus, the question arises: is it possible to construct a second-order unfolding framework that simultaneously makes full use of these optimization states?}
\begin{figure}[t]
    \centering
    \includegraphics[width=\columnwidth]{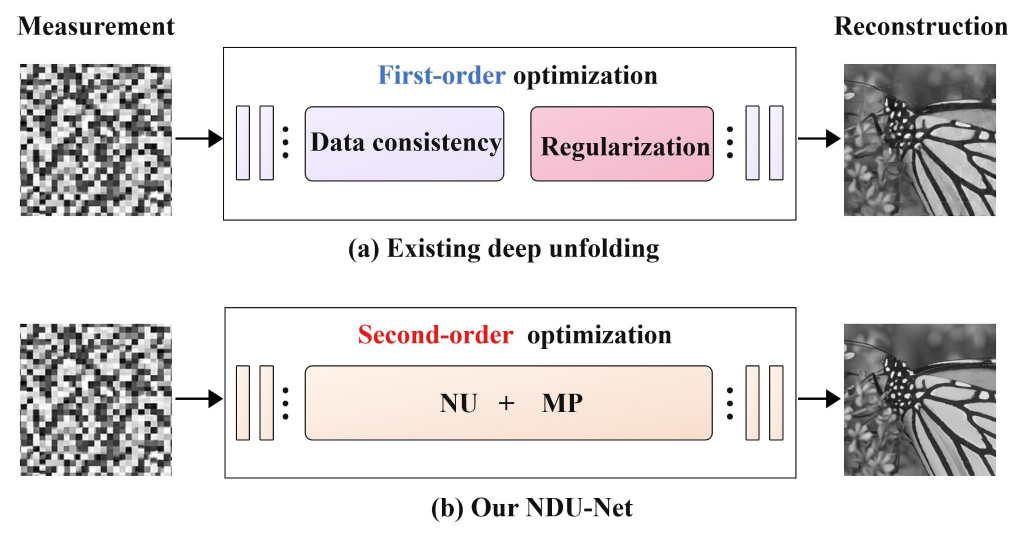}
    \caption{Comparison between typical first-order deep unfolding networks and our second-order NDU-Net.}
    \label{fig:second_order_optimization}
\end{figure}

To fill this gap, we propose a Newton deep unfolding network (NDU-Net) for CS reconstruction. As shown in Fig.~\ref{fig:set11_time_scatter}, NDU-Net achieves the highest PSNR among the compared methods on Set11 at CS ratio $=0.25$, while maintaining a moderate reconstruction time. As illustrated in Fig.~\ref{fig:second_order_optimization}, NDU-Net differs from existing first-order unfolding methods by introducing second-order optimization through the coupled Newton update (NU) and multi-scale prior (MP) modules. Compared with the existing unfolding methods, the main contributions of our NDU-Net are summarized as follows.
\begin{itemize}
\item We design a lightweight NU module for CS reconstruction. It uses only a three-layer convolutional network to estimate Newton-type update directions and generate optimization states that characterize the reconstruction status with respect to the measurement model.
\item We develop an efficient MP module for image prior restoration. It extracts local, middle-scale, and large-scale features to capture fine textures and broader structural information during stage-wise reconstruction.
\item We introduce Newton guidance within the MP module by encoding the optimization states generated by the NU module and integrating them with the fused multi-scale features. This enables prior restoration to adapt to the current data-consistency update at each unfolding stage.
\end{itemize}

\section{Related Work}
\label{sec:related_work}

\subsection{Second-Order Optimization}

Second-order iterative methods, including Newton, quasi-Newton, and Gauss-Newton algorithms, exploit Hessian-based curvature information to provide more informative update directions than first-order gradient descent~\cite{nocedal2006numerical}. In particular, Newton-type methods compute a curvature-aware correction by solving a linear system involving the gradient and Hessian, allowing image reconstruction to better account for the measurement model. However, constructing and solving such systems is computationally expensive and can be numerically unstable for high-dimensional, ill-conditioned, or underdetermined inverse problems.

Recent studies have introduced Newton-type optimization into learning-based inverse problem solvers. DeepSN-Net designs a semi-smooth Newton-driven unfolding network for blind image restoration, and SNUM-Net extends this idea to multi-modal image super-resolution. Their results indicate that Newton-driven unfolding can improve learning efficiency and restoration accuracy while preserving algorithmic interpretability~\cite{deng2025deepsnnet,zhang2025snum}. However, how to incorporate Newton-type correction into CS reconstruction and leverage the resulting optimization states to guide learned priors remains underexplored.

\subsection{Prior Structure Design}

\begin{figure*}[t]
	\centering
	\includegraphics[width=\textwidth]{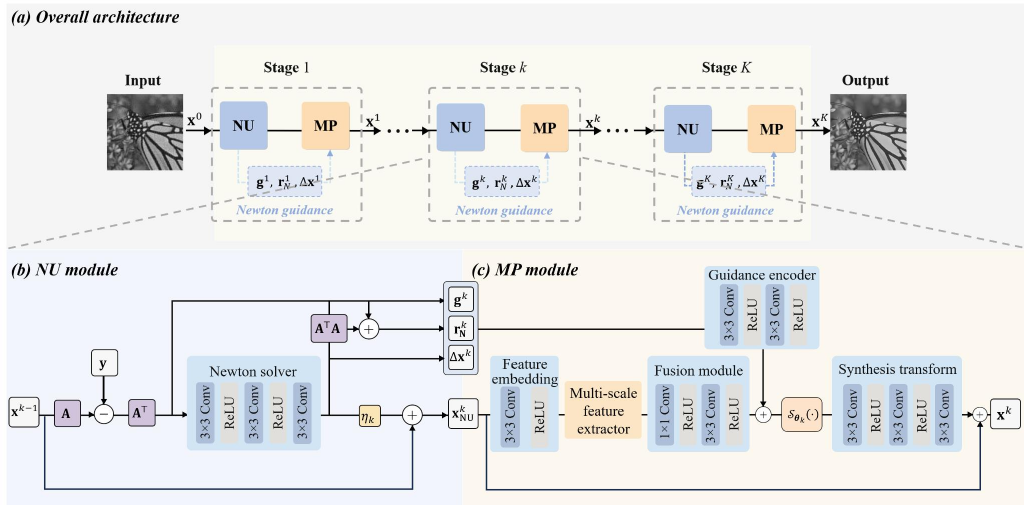}
	\caption{Overall architecture of NDU-Net. (a) The $K$-stage framework uses NU-derived optimization states $(\mathbf{g}^{k}, \mathbf{r}_{N}^{k}, \Delta\mathbf{x}^{k})$ to guide the MP module at each stage. (b) The NU module performs Newton-type data-consistency correction. (c) The MP module fuses Newton guidance with multi-scale features.}
	\label{fig:overall}
\end{figure*}

Besides the design of optimization updates, learned prior modeling is another key factor in deep unfolding networks for CS reconstruction. Early unfolding methods often replace the proximal operator or regularization step with convolutional modules, improving the representation ability of hand-crafted priors while preserving the iterative structure of optimization algorithms~\cite{zhang2018istanet,zhang2021ampnet}. However, convolution-based priors are mainly local and may be insufficient to capture long-range dependencies and complex nonlocal structures.

Recent CS unfolding methods have explored more expressive prior modules within each unfolding stage. Attention-based, multi-scale, hierarchical, and history-aware designs improve global context modeling, multi-resolution restoration, and stage-wise information reuse~\cite{song2023octuf,cui2022fhdun,guo2024cppnet,guo2025usbnet}. State-space and Mamba-based modules have also been introduced to balance long-range dependency modeling and computational efficiency~\cite{yang2026mambacs}. These developments show that learned priors have evolved from simple denoising or proximal mapping networks into structure-aware restoration components. Nevertheless, most existing prior modules are still mainly driven by image features or first-order optimization cues, while richer optimization states produced by Newton-type updates remain underexplored for guiding prior restoration.

\section{Proposed Method}
\label{sec:proposed_method}

\subsection{Overall Architecture}

The overall framework of NDU-Net is illustrated in Fig.~\ref{fig:overall}. Here, NDU-Net unfolds the CS reconstruction process into $K$ stages, with each stage comprising two coupled components: the NU module and the MP module. Specifically, the NU module estimates a Newton-type update according to the sensing model and produces optimization states, while the MP module restores image structures by exploiting both the intermediate reconstruction and the optimization states generated by the NU module.

Given the compressed measurement $\mathbf{y}$ and the sensing matrix $\mathbf{A}$, the network first obtains an initial reconstruction
\begin{equation}
	\mathbf{x}^{0}
	=
	\mathbf{A}^{\top}\mathbf{y}.
\end{equation}
Starting from $\mathbf{x}^{0}$, NDU-Net progressively refines the reconstruction through $K$ unfolding stages. At the $k$-th stage, the NU module first produces an intermediate reconstruction $\mathbf{x}_{\mathrm{NU}}^{k}$ together with optimization states that describe the current data-consistency correction. The MP module then takes $\mathbf{x}_{\mathrm{NU}}^{k}$ as the main reconstruction input and uses these optimization states to guide prior restoration. The output of the $k$-th stage is obtained by adding the learned prior correction to the intermediate reconstruction
\begin{equation}
	\mathbf{x}^{k+1}
	=
	\mathbf{x}_{\mathrm{NU}}^{k}
	+
	\mathbf{r}_{\mathrm{prior}}^{k}.
\end{equation}
Here, $\mathbf{r}_{\mathrm{prior}}^{k}$ denotes the residual correction produced by the MP module. This stage-wise design connects data-consistency correction with learned prior restoration while preserving the interpretability of deep unfolding.

\subsection{NU Module}

As shown in Fig.~\ref{fig:overall}(b), the NU module corresponds to the approximate Newton data-consistency update within each unfolding stage. Its goal is to incorporate second-order optimization information based on the current reconstruction $\mathbf{x}^{k}$ and produce an intermediate reconstruction $\mathbf{x}_{\mathrm{NU}}^{k}$.

For the $k$-th stage, the NU module operates on the data-consistency term $f(\mathbf{x})$ defined in Eq.~\eqref{eq:data_term}. Its first-order gradient at $\mathbf{x}^{k}$ is given by
\begin{equation}
	\mathbf{g}^{k}
	=
	\nabla f(\mathbf{x}^{k})
	=
	\mathbf{A}^{\top}
	(
	\mathbf{A}\mathbf{x}^{k}
	-
	\mathbf{y}
	).
\end{equation}
The Hessian matrix of this term is 
\begin{equation}
	\nabla^{2}f(\mathbf{x}^{k})
	=
	\mathbf{A}^{\top}\mathbf{A},
\end{equation}
which captures the second-order curvature induced by the sensing operator $\mathbf{A}$. Furthermore, a Newton direction $\Delta\mathbf{x}^{k}$ satisfies the following linear system
\begin{equation}
	\nabla^{2}f(\mathbf{x}^{k})
	\Delta\mathbf{x}^{k}
	+
	\nabla f(\mathbf{x}^{k})
	=
	\mathbf{0},
\end{equation}
which is equivalent to
\begin{equation}
	\mathbf{A}^{\top}\mathbf{A}
	\Delta\mathbf{x}^{k}
	+
	\mathbf{g}^{k}
	=
	\mathbf{0}.
\end{equation}

However, in general linear inverse problems, $\mathbf{A}$ is often ill-posed or underdetermined, making $\mathbf{A}^{\top}\mathbf{A}$ rank-deficient or ill-conditioned. Directly solving this Newton system is computationally expensive and numerically unstable, because it would require explicit matrix inversion or iterative linear solvers such as conjugate gradient for a high-dimensional Hessian-related system at each unfolding stage. Therefore, the NU module employs the Newton solver depicted in Fig.~\ref{fig:overall}(b), which comprises three $3\times3$ convolution layers and two ReLU activations, to approximate the Newton direction
\begin{equation}
	\Delta\mathbf{x}^{k}
	=
	\mathcal{N}^{k}
	(
	\mathbf{g}^{k}
	),
\end{equation}
where $\mathcal{N}^{k}(\cdot)$ denotes the Newton solver at the $k$-th stage. It learns a mapping from $\mathbf{g}^{k}$ to $\Delta\mathbf{x}^{k}$ to produce a data-consistency update direction. Rather than explicitly solving the Newton linear system, the learned direction is later regularized by a Newton residual that measures its consistency with the Newton equation associated with $\mathbf{A}^{\top}\mathbf{A}$.

After obtaining the approximate Newton direction, the NU module performs a data-consistency update with a learnable step size $\eta_{k}$ as 
\begin{equation}
	\mathbf{x}_{\mathrm{NU}}^{k}
	=
	\mathbf{x}^{k}
	+
	\eta_{k}
	\Delta\mathbf{x}^{k}.
\end{equation}
Here, $\eta_{k}$ controls the update magnitude at the $k$-th stage, and $\mathbf{x}_{\mathrm{NU}}^{k}$ denotes the intermediate reconstruction after the approximate Newton update.

To further enforce consistency between the estimated update direction and the ideal Newton equation, we define the Newton residual as
\begin{equation}
	\mathbf{r}_{N}^{k}
	=
	\mathbf{A}^{\top}
	\mathbf{A}
	\Delta\mathbf{x}^{k}
	+
	\mathbf{g}^{k}.
	\label{eq:newton_residual}
\end{equation}
A smaller norm of $\mathbf{r}_{N}^{k}$ indicates that the estimated direction $\Delta\mathbf{x}^{k}$ more closely satisfies the Newton equation. Therefore, $\mathbf{r}_{N}^{k}$ is used as a training constraint. In addition, the NU module passes $\mathbf{g}^{k}$, $\mathbf{r}_{N}^{k}$, and $\Delta\mathbf{x}^{k}$ to the subsequent MP module as optimization state information.

It is noticed that the U-Net solver used in a single stage of SNUM-Net contains $1.32$ M parameters~\cite{zhang2025snum}, whereas each NU module contains only $0.002626$ M parameters. This reduces the per-stage solver parameter count by $99.80\%$, demonstrating the lightweight realization of Newton-type updates in NDU-Net.

\begin{figure}[t]
    \centering
    \includegraphics[width=\columnwidth]{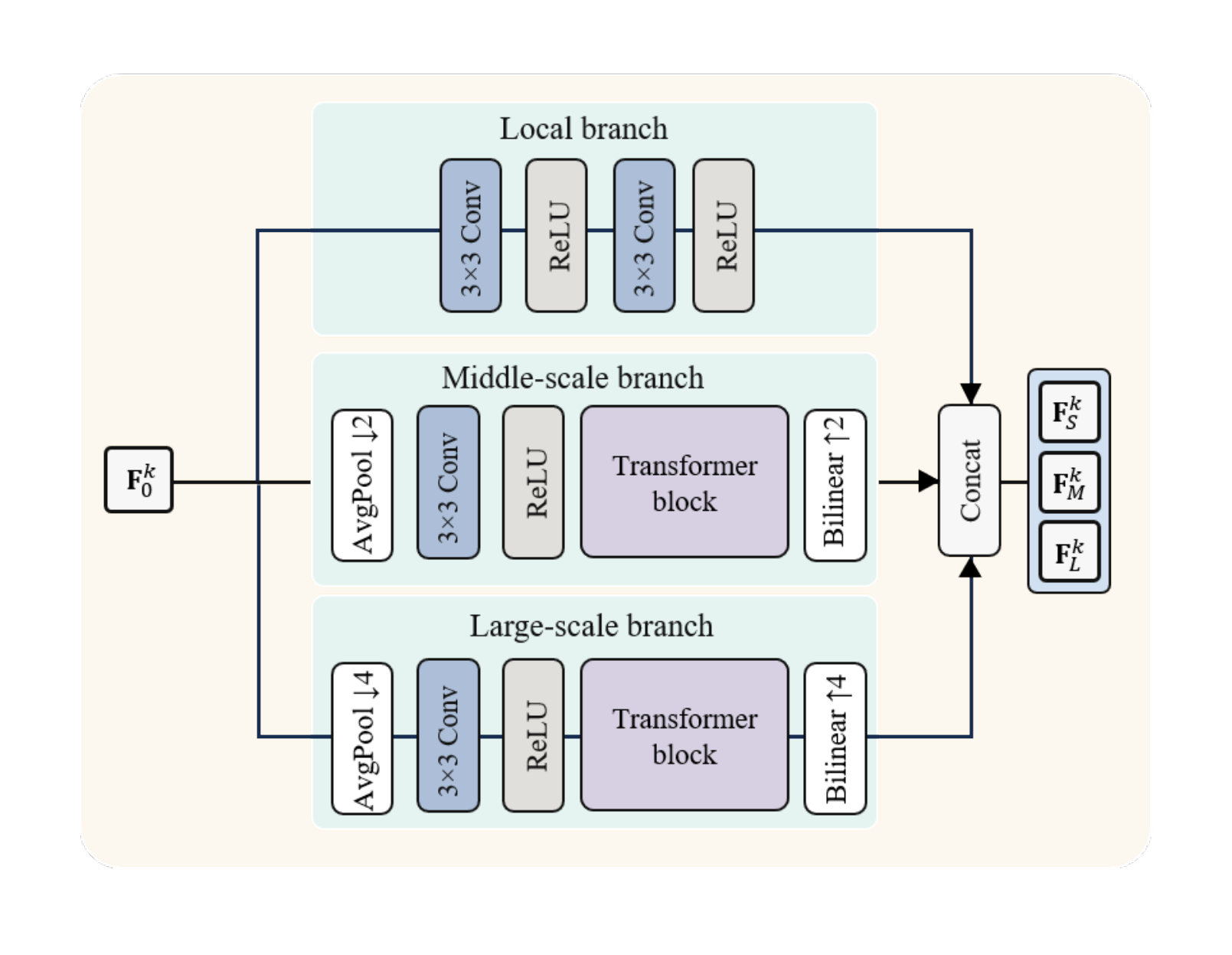}
    \caption{Multi-scale feature extractor in the MP module. Features from local, middle-scale, and large-scale branches are concatenated and fused.}
    \label{fig:multi_scale_prior_extractor}
\end{figure}

\begin{table*}[!t]
\centering
\small
\setlength{\tabcolsep}{2.0pt}
\renewcommand{\arraystretch}{1.03}
\begin{tabularx}{\textwidth}{>{\centering\arraybackslash}p{0.78in}|>{\centering\arraybackslash}p{1.62in}|YYYY|Y}
\toprule
\multirow{2}{*}{Datasets} & \multirow{2}{*}{Methods} & \multicolumn{4}{c|}{CS ratios} & \multirow{2}{*}{Average} \\
\cmidrule(lr){3-6}
& & 0.10 & 0.25 & 0.40 & 0.50 & \\
\midrule
\multirow{8}{*}{LIVE29}
& ISTA-Net+ (CVPR 2018) & 25.05/0.7104 & 29.42/0.8648 & 32.55/0.9275 & 34.55/0.9514 & 30.39/0.8635 \\
& TransCS (TIP 2022) & 27.66/0.8196 & 32.05/0.9251 & 35.51/0.9633 & 37.74/0.9767 & 33.24/0.9212 \\
& DPC-DUN (TIP 2023) & 26.83/0.7812 & 31.17/0.9008 & 34.43/0.9477 & 36.43/0.9649 & 32.22/0.8987 \\
& OCTUF (CVPR 2023) & 28.23/0.8345 & 32.69/0.9327 & 36.19/0.9671 & \textcolor{secondblue}{38.52}/\textcolor{secondblue}{0.9795} & 33.91/0.9285 \\
& CPP-Net (CVPR 2024) & 28.46/0.8414 & 32.82/0.9343 & 36.20/0.9672 & 38.37/0.9790 & 33.96/0.9305 \\
& LCNet (TMM 2025) & 28.19/0.8349 & 32.64/0.9325 & 36.08/0.9667 & 38.34/0.9791 & 33.81/0.9283 \\
& USB-Net (TIP 2025) & \textcolor{secondblue}{28.48}/\textcolor{secondblue}{0.8420} & \textcolor{secondblue}{32.87}/\textcolor{secondblue}{0.9348} & \textcolor{secondblue}{36.29}/\textcolor{secondblue}{0.9676} & 38.46/0.9793 & \textcolor{secondblue}{34.03}/\textcolor{secondblue}{0.9309} \\
& NDU-Net (ours) & \textcolor{red}{\textbf{29.46}}/\textcolor{red}{\textbf{0.8487}} & \textcolor{red}{\textbf{34.43}}/\textcolor{red}{\textbf{0.9409}} & \textcolor{red}{\textbf{38.24}}/\textcolor{red}{\textbf{0.9709}} & \textcolor{red}{\textbf{40.46}}/\textcolor{red}{\textbf{0.9807}} & \textcolor{red}{\textbf{35.65}}/\textcolor{red}{\textbf{0.9353}} \\
\midrule
\multirow{8}{*}{OST300}
& ISTA-Net+ (CVPR 2018) & 24.78/0.6896 & 28.53/0.8433 & 31.34/0.9116 & 33.20/0.9396 & 29.46/0.8460 \\
& TransCS (TIP 2022) & 27.31/0.8019 & 31.07/0.9096 & 34.17/0.9534 & 36.25/0.9701 & 32.20/0.9088 \\
& DPC-DUN (TIP 2023) & 26.25/0.7561 & 29.93/0.8792 & 32.81/0.9335 & 34.69/0.9554 & 30.92/0.8811 \\
& OCTUF (CVPR 2023) & 27.77/0.8148 & 31.60/0.9175 & 34.61/0.9572 & 36.69/0.9726 & 32.67/0.9155 \\
& CPP-Net (CVPR 2024) & 27.93/0.8207 & 31.67/0.9185 & 34.66/0.9573 & 36.68/0.9724 & 32.74/0.9172 \\
& LCNet (TMM 2025) & 27.73/0.8151 & 31.53/0.9169 & 34.53/0.9568 & 36.60/0.9723 & 32.60/0.9153 \\
& USB-Net (TIP 2025) & \textcolor{secondblue}{27.95}/\textcolor{secondblue}{0.8214} & \textcolor{secondblue}{31.71}/\textcolor{secondblue}{0.9191} & \textcolor{secondblue}{34.71}/\textcolor{secondblue}{0.9578} & \textcolor{secondblue}{36.71}/\textcolor{secondblue}{0.9726} & \textcolor{secondblue}{32.77}/\textcolor{secondblue}{0.9177} \\
& NDU-Net (ours) & \textcolor{red}{\textbf{28.76}}/\textcolor{red}{\textbf{0.8300}} & \textcolor{red}{\textbf{33.11}}/\textcolor{red}{\textbf{0.9283}} & \textcolor{red}{\textbf{36.77}}/\textcolor{red}{\textbf{0.9645}} & \textcolor{red}{\textbf{39.30}}/\textcolor{red}{\textbf{0.9775}} & \textcolor{red}{\textbf{34.49}}/\textcolor{red}{\textbf{0.9251}} \\
\midrule
\multirow{8}{*}{Set11}
& ISTA-Net+ (CVPR 2018) & 26.49/0.8036 & 32.44/0.9237 & 36.02/0.9579 & 38.07/0.9706 & 33.26/0.9140 \\
& TransCS (TIP 2022) & 29.54/0.8877 & 35.06/0.9548 & 38.46/0.9737 & 40.50/0.9815 & 35.89/0.9494 \\
& DPC-DUN (TIP 2023) & 29.40/0.8798 & 34.69/0.9482 & 37.98/0.9694 & 39.84/0.9778 & 35.48/0.9438 \\
& OCTUF (CVPR 2023) & 30.70/0.9030 & 36.10/0.9604 & 39.41/0.9773 & 41.34/0.9838 & 36.89/0.9561 \\
& CPP-Net (CVPR 2024) & 31.27/\textcolor{secondblue}{0.9135} & 36.35/0.9631 & 39.53/0.9781 & 41.39/0.9842 & 37.14/0.9597 \\
& LCNet (TMM 2025) & 30.40/0.9010 & 35.77/0.9593 & 39.08/0.9762 & 41.02/0.9831 & 36.57/0.9549 \\
& USB-Net (TIP 2025) & \textcolor{secondblue}{31.31}/\textcolor{red}{\textbf{0.9149}} & \textcolor{secondblue}{36.42}/\textcolor{secondblue}{0.9632} & \textcolor{secondblue}{39.64}/\textcolor{secondblue}{0.9785} & \textcolor{secondblue}{41.47}/\textcolor{secondblue}{0.9843} & \textcolor{secondblue}{37.21}/\textcolor{secondblue}{0.9602} \\
& NDU-Net (ours) & \textcolor{red}{\textbf{31.82}}/\textcolor{secondblue}{0.9135} & \textcolor{red}{\textbf{37.34}}/\textcolor{red}{\textbf{0.9657}} & \textcolor{red}{\textbf{40.68}}/\textcolor{red}{\textbf{0.9799}} & \textcolor{red}{\textbf{42.59}}/\textcolor{red}{\textbf{0.9852}} & \textcolor{red}{\textbf{38.11}}/\textcolor{red}{\textbf{0.9611}} \\
\midrule
\multirow{8}{*}{BSD68}
& ISTA-Net+ (CVPR 2018) & 27.48/0.7658 & 32.40/0.9019 & 35.90/0.9515 & 38.05/0.9688 & 33.46/0.8970 \\
& TransCS (TIP 2022) & 30.43/0.8654 & 35.61/0.9520 & 39.64/0.9794 & 42.17/0.9880 & 36.96/0.9462 \\
& DPC-DUN (TIP 2023) & 29.20/0.8228 & 34.11/0.9289 & 37.89/0.9671 & 40.27/0.9800 & 35.37/0.9247 \\
& OCTUF (CVPR 2023) & 31.04/0.8743 & 36.23/0.9564 & 40.25/0.9815 & \textcolor{secondblue}{42.86}/\textcolor{secondblue}{0.9894} & 37.60/0.9504 \\
& CPP-Net (CVPR 2024) & 31.26/0.8791 & 36.27/0.9571 & 40.20/0.9814 & 42.71/0.9891 & 37.61/0.9517 \\
& LCNet (TMM 2025) & 30.91/0.8735 & 36.10/0.9558 & 40.14/0.9813 & 42.71/0.9893 & 37.47/0.9500 \\
& USB-Net (TIP 2025) & \textcolor{secondblue}{31.26}/\textcolor{secondblue}{0.8794} & \textcolor{secondblue}{36.31}/\textcolor{secondblue}{0.9574} & \textcolor{secondblue}{40.26}/\textcolor{secondblue}{0.9815} & 42.75/0.9892 & \textcolor{secondblue}{37.65}/\textcolor{secondblue}{0.9519} \\
& NDU-Net (ours) & \textcolor{red}{\textbf{31.94}}/\textcolor{red}{\textbf{0.8797}} & \textcolor{red}{\textbf{37.61}}/\textcolor{red}{\textbf{0.9599}} & \textcolor{red}{\textbf{41.84}}/\textcolor{red}{\textbf{0.9828}} & \textcolor{red}{\textbf{44.37}}/\textcolor{red}{\textbf{0.9897}} & \textcolor{red}{\textbf{38.94}}/\textcolor{red}{\textbf{0.9530}} \\
\bottomrule
\end{tabularx}
\caption{Average PSNR/SSIM comparisons on four benchmark datasets under different CS ratios. Red bold and blue values denote the best and second-best results, respectively.}
\label{tab:cs_compare}
\end{table*}

\begin{figure*}[!t]
    \centering
    \includegraphics[width=\textwidth]{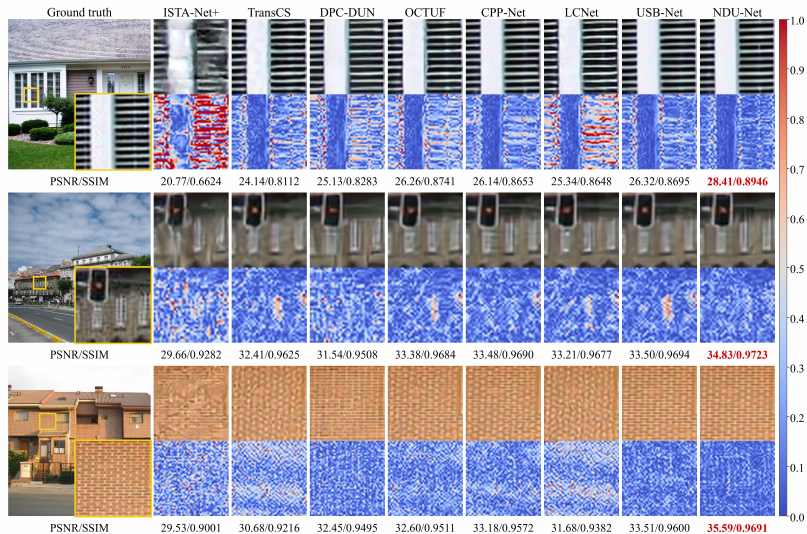}
    \caption{Visual comparisons of reconstructed images, including local detail enlargements and error heatmaps. From top to bottom, the CS ratios are $0.10$, $0.25$, and $0.40$, respectively. }
    \label{fig:qualitative_comparison}
\end{figure*}

\begin{figure*}[t]
    \centering
    \includegraphics[width=0.75\textwidth]{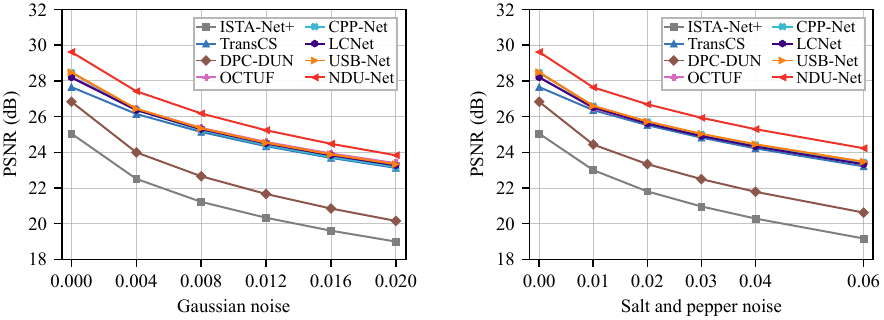}
    \caption{Robustness comparisons on LIVE29 at CS ratio $=0.10$, where the left corresponds to Gaussian noise and the right corresponds to  salt-and-pepper noise.}
    \label{fig:noise_robustness}
\end{figure*}

\subsection{MP Module}

As shown in Fig.~\ref{fig:overall}(c), the MP module takes $\mathbf{x}_{\mathrm{NU}}^{k}$ as the main reconstruction input and uses the optimization state variables $(\mathbf{g}^{k}, \mathbf{r}_{N}^{k}, \Delta\mathbf{x}^{k})$ to guide multi-scale feature restoration. The output of the MP module is a residual correction term $\mathbf{r}_{\mathrm{prior}}^{k}$, which is added to $\mathbf{x}_{\mathrm{NU}}^{k}$ to obtain the output of the current unfolding stage.

First, to form a compact guidance representation, the optimization state variables are concatenated along the channel dimension and encoded by a guidance encoder
\begin{equation}
	\mathbf{G}^{k}
	=
	\mathcal{E}^{k}
	(
	\mathbf{g}^{k},
	\mathbf{r}_{N}^{k},
	\Delta\mathbf{x}^{k}
	),
\end{equation}
where $\mathcal{E}^{k}(\cdot)$ consists of two $3\times3$ convolution layers with ReLU activations. The encoded feature $\mathbf{G}^{k}$ provides Newton-guided optimization information for the subsequent prior restoration process.

Meanwhile, $\mathbf{x}_{\mathrm{NU}}^{k}$ is processed by the multi-scale feature extractor illustrated in Fig.~\ref{fig:multi_scale_prior_extractor}, which is given by
\begin{equation}
	\mathbf{F}_{\mathrm{MS}}^{k}
	=
	\mathcal{P}^{k}
	(
	\mathbf{x}_{\mathrm{NU}}^{k}
	),
\end{equation}
where $\mathcal{P}^{k}(\cdot)$ comprises shallow embedding, multi-scale feature extraction, and feature fusion. First, $\mathbf{x}_{\mathrm{NU}}^{k}$ is mapped to a shallow feature $\mathbf{F}_{0}^{k}$ through a $3\times3$ convolution followed by ReLU. Next, three parallel branches process $\mathbf{F}_{0}^{k}$ to produce the local-, middle-, and large-scale features $\mathbf{F}_{S}^{k}$, $\mathbf{F}_{M}^{k}$, and $\mathbf{F}_{L}^{k}$, respectively. The local branch preserves the original spatial resolution and applies two $3\times3$ convolution layers with ReLU activations to capture fine-grained structures. Following hierarchical multi-scale designs~\cite{liu2021swin,cui2022fhdun,song2023octuf,guo2024cppnet}, the middle-scale and large-scale branches downsample $\mathbf{F}_{0}^{k}$ by factors of $2$ and $4$, respectively, process the downsampled features using a $3\times3$ convolution, ReLU, and window-based Transformer blocks, and restore them to the original resolution through bilinear upsampling. Finally, $\mathbf{F}_{S}^{k}$, $\mathbf{F}_{M}^{k}$, and $\mathbf{F}_{L}^{k}$ are concatenated and fused using a $1\times1$ convolution, ReLU, a $3\times3$ convolution, and ReLU to produce $\mathbf{F}_{\mathrm{MS}}^{k}$.

In particular, the window-based Transformer block used in the middle-scale and large-scale branches operates on local windows. Given an input feature, the feature is partitioned into non-overlapping local windows. Within each window, standard multi-head self-attention and a feed-forward network are applied with residual connections and layer normalization~\cite{vaswani2017attention,liu2021swin}. If the spatial size is not divisible by the window size, reflection padding is used before window partition and the padded regions are cropped after window reversal. This window-based design captures contextual dependencies within each scale while avoiding the cost of global self-attention.

The fused multi-scale feature is then combined with the Newton-guided feature by direct addition as follows
\begin{equation}
	\mathbf{F}_{\mathrm{guided}}^{k}
	=
	\mathbf{F}_{\mathrm{MS}}^{k}
	+
	\mathbf{G}^{k}.
\end{equation}
This operation injects the optimization state information into the learned prior feature.

Motivated by sparse regularization for retaining informative features and suppressing irrelevant noise~\cite{xiu2025bisparse}, we apply a learnable soft-thresholding operation, following optimization-inspired sparse prior modeling in ISTA-based unfolding networks~\cite{zhang2018istanet}, to enhance feature-domain sparsity
\begin{equation}
	\mathbf{F}_{\mathrm{shrink}}^{k}
	=
	\mathcal{S}_{\boldsymbol{\theta}_{k}}
	(
	\mathbf{F}_{\mathrm{guided}}^{k}
	),
\end{equation}
where the soft-thresholding operator $\mathcal{S}_{\boldsymbol{\theta}_{k}}(\cdot)$ is defined as
\begin{equation}
	\mathcal{S}_{\boldsymbol{\theta}_{k}}(\mathbf{z})
	=
	\operatorname{sign}(\mathbf{z})
	\odot
	\max
	(
	|
	\mathbf{z}
	|
	-
	\boldsymbol{\theta}_{k},
	0
	),
\end{equation}
where $\boldsymbol{\theta}_{k}$ is a learnable threshold and $\odot$ denotes element-wise multiplication. Finally, a synthesis transform maps the shrunk feature back to the image domain
\begin{equation}
	\mathbf{r}_{\mathrm{prior}}^{k}
	=
	\mathcal{H}^{k}
	(
	\mathbf{F}_{\mathrm{shrink}}^{k}
	),
\end{equation}
where $\mathcal{H}^{k}(\cdot)$ is composed of three $3\times3$ convolution layers with ReLU activations between adjacent convolutions. The output of the $k$-th stage is then obtained by
\begin{equation}
	\mathbf{x}^{k+1}
	=
	\mathbf{x}_{\mathrm{NU}}^{k}
	+
	\mathbf{r}_{\mathrm{prior}}^{k}.
\end{equation}

We would like to point out that CPP-Net emphasizes multi-scale image-feature extraction and fusion through a dual-path block~\cite{guo2024cppnet}, while existing Newton-type unfolding methods for other inverse problems incorporate second-order corrections without using the resulting optimization states to guide learned priors~\cite{deng2025deepsnnet,zhang2025snum}. In contrast, the proposed MP module conditions the fused multi-scale feature on the Newton-derived optimization states $(\mathbf{g}^{k}, \mathbf{r}_{N}^{k}, \Delta\mathbf{x}^{k})$. This explicit Newton guidance aligns prior correction with the current data-consistency status and enables it to adapt across unfolding stages.

\subsection{Loss Function}
\label{subsec:loss_function}

The training objective of NDU-Net consists of a reconstruction loss and a Newton residual constraint. In detail, the reconstruction loss is applied to the final output $\mathbf{x}^{K}$ and is defined as the mean absolute error
\begin{equation}
    \mathcal{L}_{\mathrm{Rec}}
    =
    \frac{1}{N}
    \|
    \mathbf{x}^{K}
    -
    \mathbf{x}
    \|_{1},
\end{equation}

To regularize the approximate Newton update, we introduce a Newton residual loss over all unfolding stages
\begin{equation}
    \mathcal{L}_{\mathrm{Newton}}
    =
    \frac{1}{K}
    \left(
    \sum_{k=0}^{K-1}
    \left(
    \frac{1}{N}
    \|
    \mathbf{r}_{N}^{k}
    \|_{1}
    \right)
    \right),
\end{equation}
where $\mathbf{r}_{N}^{k}$ is the Newton residual defined in Eq.~\eqref{eq:newton_residual}. The overall training objective is given by
\begin{equation}
    \mathcal{L}
    =
    \mathcal{L}_{\mathrm{Rec}}
    +
    \beta
    \mathcal{L}_{\mathrm{Newton}},
\end{equation}
where $\beta > 0$ controls the weight of the Newton residual constraint.

\section{Experiments}
\label{sec:experiments}

\subsection{Implementation Details}
\label{subsec:implementation_details}

We train the proposed NDU-Net on images from the Waterloo exploration database (WED)~\cite{ma2017waterloo} using randomly cropped $128 \times 128$ patches. We utilize four benchmark datasets for evaluation, including LIVE29~\cite{sheikh2006live}, OST300~\cite{wang2018esrgan}, Set11~\cite{kulkarni2016reconnet}, and BSD68~\cite{martin2001database}, and report PSNR and SSIM~\cite{wang2004image}. For each CS ratio of $0.10$, $0.25$, $0.40$, and $0.50$, we independently train an NDU-Net model. We apply Adam with an initial learning rate of $1\times10^{-4}$, a batch size of 16, and 200 training epochs. The loss weight $\beta$ is initialized to $0.01$ and decayed during the training process. The proposed NDU-Net is trained and evaluated on a server equipped with one NVIDIA RTX PRO 6000 GPU (96 GB) and 25 Intel Xeon Platinum 8470Q vCPUs.

\subsection{Quantitative Results}
\label{quantitative results}

As shown in Table~\ref{tab:cs_compare}, NDU-Net achieves the best PSNR across all benchmark datasets and CS ratios and the best SSIM in most settings, demonstrating consistent reconstruction performance under diverse image contents and sampling conditions. The visual comparisons in Fig.~\ref{fig:qualitative_comparison} further examine three representative enlarged regions, including the repeated horizontal stripes of window louvers, a small high-contrast sign on a building facade, and dense brick-wall textures. Compared with competing methods, NDU-Net better preserves stripe continuity and straight boundaries in the first region, retains the shape and contrast of the small sign in the second, and reconstructs more regular brick patterns in the third, while producing lower error responses. These visual results are consistent with the quantitative improvements.

\subsection{Robustness Analysis}
\label{subsec:sensitivity_to_noise}

Fig.~\ref{fig:noise_robustness} compares measurement-noise robustness on LIVE29 at the CS ratio of $0.1$ under Gaussian and salt-and-pepper noise. Although all methods degrade as the noise level increases, NDU-Net maintains the highest PSNR throughout the tested ranges. At a Gaussian noise variance of $0.02$ and a salt-and-pepper noise density of $0.06$, it achieves $23.83$ and $24.23$ dB, exceeding the best competing results by $0.44$ and $0.75$ dB, respectively. Since the optimization states are derived from corrupted measurements, this consistent advantage suggests that their integration with multi-scale features avoids substantial noise amplification.

\subsection{Ablation Studies}	
\label{subsec:ablation_studies}

\begin{table}[t]
\centering
\small
\setlength{\tabcolsep}{2.4pt}
\renewcommand{\arraystretch}{1.05}
\begin{tabularx}{\columnwidth}{@{}lYYY@{}}
\toprule
Variants & LIVE29 & Set11 & BSD68 \\
\midrule
w/o NU & 33.91/0.9307 & 36.92/0.9625 & 36.80/0.9505 \\
w/o MP & 33.16/0.9257 & 35.70/0.9545 & 36.21/0.9483 \\
Full model & \textbf{34.43/0.9409} & \textbf{37.34/0.9657} & \textbf{37.61/0.9599} \\
\bottomrule
\end{tabularx}
\caption{Ablation studies of the NU and MP modules at CS ratio $=0.25$.}
\label{tab:ablation_components}
\end{table}

To verify the effectiveness of the proposed modules, we conduct ablation studies at the CS ratio of $0.25$. Table~\ref{tab:ablation_components} compares three variants on LIVE29, Set11, and BSD68. Replacing the NU module with a first-order data-consistency update reduces PSNR by $0.52$, $0.42$, and $0.81$ dB on the three datasets, respectively, indicating the consistent benefit of the Newton-type update. Replacing the MP module with a CNN prior causes larger PSNR drops of $1.27$, $1.64$, and $1.40$ dB, respectively, together with lower SSIM, demonstrating the importance of the MP module for prior restoration. Accordingly, the full NDU-Net achieves the best PSNR and SSIM on all three datasets.

We further analyze the influence of the number of unfolding stages on Set11 at the CS ratio of $0.25$. As shown in Table~\ref{tab:layer_ablation}, increasing the number of stages from 1 to 2 yields the largest PSNR gain of $2.29$ dB, while the improvements gradually diminish in later stages. The performance peaks at 6 stages with $37.34$ dB PSNR and $0.9657$ SSIM, corresponding to gains of $3.90$ dB and $0.0306$ over the 1-stage model. Further extending the unfolding process to 7 stages raises the parameter count from $5.989$ M to $6.987$ M, yet slightly decreases PSNR and SSIM by $0.03$ dB and $0.0003$, respectively. Therefore, we set $K = 6$ in all experiments to balance reconstruction quality and model size.

\begin{table}[t]
\centering
\small
\setlength{\tabcolsep}{4pt}
\renewcommand{\arraystretch}{1.05}
\begin{tabularx}{\columnwidth}{YYYY}
\toprule
Layers & PSNR & SSIM & Params \\
\midrule
1 & 33.44 & 0.9351 & 1.001 \\
2 & 35.73 & 0.9550 & 1.999 \\
3 & 36.67 & 0.9612 & 2.996 \\
4 & 37.02 & 0.9635 & 3.994 \\
5 & 37.29 & 0.9653 & 4.991 \\
6 & \textbf{37.34} & \textbf{0.9657} & 5.989 \\
7 & 37.31 & 0.9654 & 6.987 \\
\bottomrule
\end{tabularx}
\caption{Ablation studies of the number of unfolding stages on Set11 at CS ratio $=0.25$.}
\label{tab:layer_ablation}
\end{table}

\section{Conclusion}
\label{sec:conclusion}

In this paper, we have developed a novel CS reconstruction method called NDU-Net. The core idea is to introduce second-order optimization into CS deep unfolding through a lightweight NU module and use the resulting Newton-derived optimization states to guide multi-scale feature restoration in the MP module. This design establishes an explicit connection between data-consistency correction and learned image priors while preserving the interpretability of the unfolding process. Extensive experiments across multiple benchmark datasets and CS ratios demonstrate that NDU-Net achieves superior reconstruction performance and strong robustness .

\bibliography{mybib}

\end{document}